\documentclass[11pt,a4paper]{article}
\usepackage[hyperref]{emnlp2018}
\usepackage{times}
\usepackage{latexsym}
\usepackage{url}

\usepackage{amsmath}
\usepackage{amssymb,amsmath,amsthm,enumitem}
\usepackage{comment}

\usepackage{graphicx}
\usepackage[english]{babel}
\usepackage{booktabs}
\usepackage{multicol}
\usepackage{multirow}

\newcommand{\Amat}[0]{{{\bf A}}}

\newcommand{\Wmat}[0]{{{\bf W}}}
\newcommand{\Xmat}[0]{{{\bf X}}}

\newcommand{\cv}[0]{{\boldsymbol{c}}}

\newcommand{\hv}[0]{{\boldsymbol{h}}}

\newcommand{\mv}[0]{{\boldsymbol{m}}}

\newcommand{\xv}{\boldsymbol{x}}

\newcommand{\Lcal}{\mathcal{L}}

\aclfinalcopy 
\def\aclpaperid{190} 

\title{Improved Semantic-Aware Network Embedding\\with Fine-Grained Word Alignment}

\author{Dinghan Shen, ~Xinyuan Zhang, ~Ricardo Henao,  Lawrence Carin
	\smallskip 
	\\
	\smallskip 
	Department of Electrical and Computer Engineering \\
	Duke University, Durham, NC, USA \\
	{\tt \{dinghan.shen, xy.zhang, r.henao, lcarin\}@duke.edu} }

\date{}

\begin{document}
\maketitle
\begin{abstract}
	Network embeddings, which learn low-dimensional representations for each vertex in a large-scale network, have received considerable attention in recent years.
	For a wide range of applications, vertices in a network are typically accompanied by rich textual information such as user profiles, paper abstracts, \emph{etc}.
	We propose to incorporate semantic features into network embeddings by matching important words between text sequences for all pairs of vertices.
	We introduce a \emph{word-by-word} alignment framework that measures the compatibility of embeddings between word pairs, and then adaptively accumulates these alignment features with a simple yet effective aggregation function.
	In experiments, we evaluate the proposed framework on three real-world benchmarks for downstream tasks, including link prediction and multi-label vertex classification.
	Results demonstrate that our model outperforms state-of-the-art network embedding methods by a large margin.
\end{abstract}

\section{Introduction}
%
%
Networks are ubiquitous, with prominent examples including social networks (\emph{e.g.}, Facebook, Twitter) or citation networks of research papers (\emph{e.g.}, arXiv).
When analyzing data from these real-world networks, 
traditional methods often represent vertices (nodes) as one-hot representations (containing the connectivity information of each vertex with respect to all other vertices), usually suffering from issues related to the inherent sparsity of large-scale networks.
This results in models that are not able to fully capture the relationships between vertices of the network \cite{perozzi2014deepwalk, tu2016max}.
Alternatively, network embedding (\emph{i.e.}, network representation learning) has been considered, representing each vertex of a network with a \emph{low-dimensional vector} that preserves information on its similarity relative to other vertices.
This approach has attracted considerable attention in recent years \cite{tang2009relational, perozzi2014deepwalk, tang2015line, grover2016node2vec, wang2016structural, chen2016incorporate, wang2017community, zhang2018diffusion}.

Traditional network embedding approaches focus primarily on learning representations of vertices that preserve local structure, as well as internal structural properties of the network.
For instance, Isomap \cite{tenenbaum2000global}, LINE \cite{tang2015line}, and Grarep \cite{cao2015grarep} were proposed to preserve first-, second-, and higher-order proximity between nodes, respectively.
DeepWalk \cite{perozzi2014deepwalk}, which learns vertex representations from random-walk sequences, similarly, only takes into account structural information of the network.
However, in real-world networks, vertices usually contain rich textual information (\emph{e.g.}, user profiles in Facebook, paper abstracts in arXiv, user-generated content on Twitter, \emph{etc}.), which may be leveraged effectively for learning more informative embeddings.

\begin{figure}[!t]
	\centering
	\includegraphics[width=.3\textwidth]{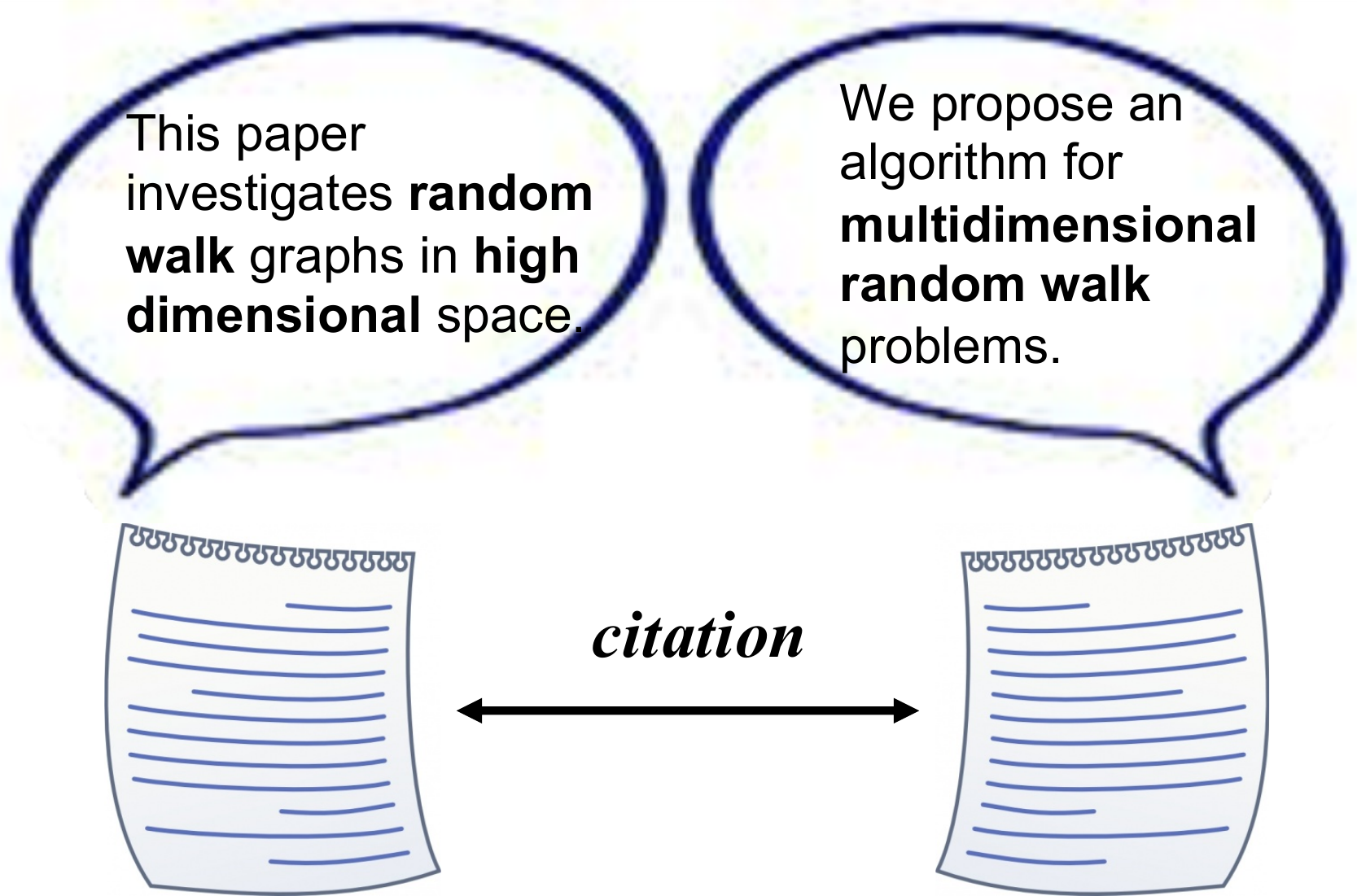}
	\caption{Example of the text information (abstracts) associated to two papers in a citation network. Key words for matching are highlighted.}
	\label{fig:example}
	\vspace{-3mm}
\end{figure}

To address this opportunity, \citet{yang2015network} proposed text-associated DeepWalk, to incorporate textual information into the vectorial representations of vertices (embeddings).
\citet{sun2016general} employed deep recurrent neural networks to integrate the information from vertex-associated text into network representations.
Further, \citet{tu2017cane} proposed to more effectively model the semantic relationships between vertices using a mutual attention mechanism.

Although these methods have demonstrated performance gains over structure-only network embeddings, the relationship between text sequences for a pair of vertices is accounted for solely by comparing their sentence embeddings.
However, as shown in Figure~\ref{fig:example}, to assess the similarity between two research papers, a more effective strategy would compare and align (via local-weighting) individual important words (keywords) within a pair of abstracts, while information from other words (\emph{e.g.}, stop words) that tend to be less relevant can be effectively ignored (down-weighted).
This alignment mechanism is difficult to accomplish in models where text sequences are first embedded into a common space and then compared in pairs \cite{he2016pairwise, parikh2016decomposable, Wang2016ACM, wang2017bilateral, shen2018context}.

We propose to learn a semantic-aware Network Embedding (NE) that incorporates word-level alignment features abstracted from text sequences associated with vertex pairs.
Given a pair of sentences, our model first aligns each word within one sentence with keywords from the other sentence (adaptively up-weighted via an attention mechanism), producing a set of fine-grained matching vectors.
These features are then accumulated via a simple but efficient aggregation function, obtaining the final representation for the sentence.
As a result, the word-by-word alignment features (as illustrated in Figure~\ref{fig:example}) are explicitly and effectively captured by our model.
Further, the learned network embeddings under our framework are adaptive to the specific (local) vertices that are considered, and thus are context-aware and especially suitable for downstream tasks, such as link prediction.
Moreover, since the word-by-word matching procedure introduced here is highly parallelizable and does not require any complex encoding networks, such as Long Short-Term Memory (LSTM) or Convolutional Neural Networks (CNNs), our framework requires significantly less time for training, which is attractive for large-scale network applications.

We evaluate our approach on three real-world datasets spanning distinct network-embedding-based applications: link prediction, vertex classification and visualization.
We show that the proposed word-by-word alignment mechanism efficiently incorporates textual information into the network embedding, and consistently exhibits superior performance relative to several competitive baselines.
Analyses considering the extracted word-by-word pairs further validate the effectiveness of the proposed framework.
%
%
%
%

\section{Proposed Methods}
\subsection{Problem Definition}
A network (graph) is defined as $\boldsymbol{G} = \{\boldsymbol{V},\boldsymbol{E}\}$, where $\boldsymbol{V}$ and $\boldsymbol{E}$
denote the set of $N$ vertices (nodes) and edges, respectively, where elements of $\boldsymbol{E}$ are two-element subsets of $\boldsymbol{V}$.
Here we only consider undirected networks, however, our approach (introduced below) can be readily extended to the directed case.
We also define $\boldsymbol{W}$, the symmetric $\mathbb{R}^{N \times N}$ matrix whose elements, $w_{ij}$, denote the weights associated with edges in $\boldsymbol{V}$, and $\boldsymbol{T}$, the set of text sequences assigned to each vertex.
Edges and weights contain the structural information of the network, while the text can be used to characterize the semantic properties of each vertex.
Given network $\boldsymbol{G}$, with the \emph{network embedding} we seek to encode each vertex into a low-dimensional vector $\hv$ (with dimension much smaller than $N$), while preserving structural and semantic features of $\boldsymbol{G}$.

\subsection{Framework Overview}
To incorporate both structural and semantic information into the network embeddings, we specify two types of (latent) embeddings: ($i$) $\hv_s$, the \emph{structural embedding}; and ($ii$) $\hv_t$, the \emph{textual embedding}.
Specifically, each vertex in $\boldsymbol{G}$ is encoded into a low-dimensional embedding $\hv = [\hv_s; \hv_t]$.
To learn these embeddings, we specify an objective that leverages the information from both $\boldsymbol{W}$ and $\boldsymbol{T}$, denoted as
\begin{align}\label{eq:obj}
	\Lcal = \sum_{\boldsymbol{e} \in \boldsymbol{E}} \Lcal_{\textrm{struct}}(\boldsymbol{e}) + 	\Lcal_{\textrm{text}}(\boldsymbol{e}) + \Lcal_{\textrm{joint}}(\boldsymbol{e}) \,,
\end{align}
%
where $\Lcal_{\textrm{struct}}$, $\Lcal_{\textrm{text}}$ and $\Lcal_{\textrm{joint}}$ denote structure, text, and joint structure-text training losses, respectively.
For a vertex pair $\{v_i,v_j\}$ weighted by $w_{ij}$, $\Lcal_{\textrm{struct}}(v_i, v_j)$ in \eqref{eq:obj} is defined as \cite{tang2015line}
%
\begin{align}\label{eq:str}
	\Lcal_{\textrm{struct}}(v_i, v_j) = w_{ij} \log p(\hv^i_s|\hv^j_{s}) \,,
\end{align}
where $p(\hv^i_s|\hv^j_{s})$ denotes the conditional probability between structural embeddings for vertices $\{v_i,v_j\}$.
To leverage the textual information in $\boldsymbol{T}$, similar text-specific and joint structure-text training objectives are also defined
\begin{align}\label{eq:text}
	\Lcal_{\textrm{text}}(v_i, v_j) & = w_{ij} \alpha_1 \log p(\hv^i_t|\hv^j_{t}) \,, \\
	\Lcal_{\textrm{joint}}(v_i, v_j) & = w_{ij} \alpha_2 \log p(\hv^i_t|\hv^j_{s}) \\ 
	& + w_{ij}\alpha_3 \log p(\hv^i_s|\hv^j_{t}) \,,
\end{align}
where $p(\hv^i_t|\hv^j_t)$ and $p(\hv^i_t|\hv^j_s)$ (or $p(\hv^i_s|\hv^j_t)$) denote the conditional probability for a pair of text embeddings and text embedding given structure embedding (or \emph{vice versa}), respectively, for vertices $\{v_i,v_j\}$.
Further, $\alpha_1$, $\alpha_2$ and $\alpha_3$ are hyperparameters that balance the impact of the different training-loss components.
Note that 
structural embeddings, $\hv_s$, are treated directly as parameters, while the text embeddings $\hv_t$ are learned based on the text sequences associated with vertices.

For all conditional probability terms, we follow \citet{tang2015line} and consider the second-order proximity between vertex pairs.
Thus, for vertices $\{v_i,v_j\}$, the probability of generating $\hv_i$ conditioned on $\hv_j$ may be written as
\begin{align}\label{eq:def}
	p(\hv^i|\hv^j) = \frac{\exp\left({\hv^j}^T \hv^i\right)}{\textstyle{\sum}_{k=1}^{N}\exp\left({\hv^j}^T \hv^k\right)} \,.
\end{align}
Note that \eqref{eq:def} can be applied to both structural and text embeddings in \eqref{eq:str} and \eqref{eq:text}.

Inspired by \citet{tu2017cane}, we further assume that vertices in the network play different roles depending on the vertex with which they interact.
Thus, for a given vertex, the text embedding, $\hv_t$, is adaptive (specific) to the vertex it is being conditioned on.
This type of context-aware textual embedding has demonstrated superior performance relative to context-free embeddings \cite{tu2017cane}.
In the following two sections, we describe our strategy for encoding the text sequence associated with an edge into its adaptive textual embedding, via word-by-context and word-by-word alignments.

\subsection{Word-by-Context Alignment}
We first introduce our base model, which re-weights the importance of individual words within a text sequence in the context of the edge being considered.
Consider text sequences associated with two vertices connected by an edge, denoted $\boldsymbol{t}_a$ and $\boldsymbol{t}_b$ and contained in $\boldsymbol{T}$.
Text sequences $\boldsymbol{t}_a$ and $\boldsymbol{t}_b$ are of lengths $M_a$ and $M_b$, respectively, and are represented by $\Xmat_a\in\mathbb{R}^{d\times M_a}$ and $\Xmat_b\in\mathbb{R}^{d\times M_b}$, respectively, where $d$ is the dimension of the word embedding.
Further, $\xv^{(i)}_a$ denotes the embedding of the $i$-th word in sequence $\boldsymbol{t}_a$.

Our goal is to encode text sequences $\boldsymbol{t}_a$ and $\boldsymbol{t}_b$ into \emph{counterpart-aware} vectorial representations $\hv_a$ and $\hv_b$.
%
%
%
Thus, while inferring the adaptive textual embedding for sentence $\boldsymbol{t}_a$, we propose re-weighting the importance of each word in $\boldsymbol{t}_a$ to explicitly account for its alignment with sentence $\boldsymbol{t}_b$.
The weight $\alpha_i$, corresponding to the $i$-th word in $\boldsymbol{t}_a$, is generated as:
\begin{align}\label{eqn:alpha}
\alpha_i = \frac{\exp(\tanh(\Wmat_1 \cv_b + \Wmat_2 \xv^{(i)}_a))}{\sum_{j = 1}^{M_a} \exp(\tanh(\Wmat_1 \cv_b + \Wmat_2 \xv^{(j)}_a))} \,,
\end{align}
where $\Wmat_1$ and $\Wmat_2$ are model parameters and $\cv_b = \sum_{i = 1}^{M_b} \boldsymbol{x}^b_i$ is the context vector of sequence $\boldsymbol{t}_b$, obtained by simply averaging over all the word embeddings in the sequence, similar to fastText \cite{joulin2016bag}.
Further, the \emph{word-by-context} embedding for sequence $\boldsymbol{t}_a$ is obtained by taking the weighted average over all word embeddings
\begin{align}\label{eqn:weighted}
	\boldsymbol{h}_a = \textstyle{\sum}_{i = 1}^{M_a} \alpha_i \xv^{(i)}_a \,.
\end{align}
Intuitively, $\alpha_i$ may be understood as the relevance score between the $i$th word in $\boldsymbol{t}_a$ and sequence $\boldsymbol{t}_b$.
Specifically, keywords within $\boldsymbol{t}_a$, in the context of $\boldsymbol{t}_b$, should be assigned larger weights, while less important words will be correspondingly down-weighted.
Similarly, $\boldsymbol{h}_b$ is encoded as a weighted embedding using \eqref{eqn:alpha} and \eqref{eqn:weighted}.

\begin{figure}[!t]
	\centering
	\includegraphics[width=0.45\textwidth]{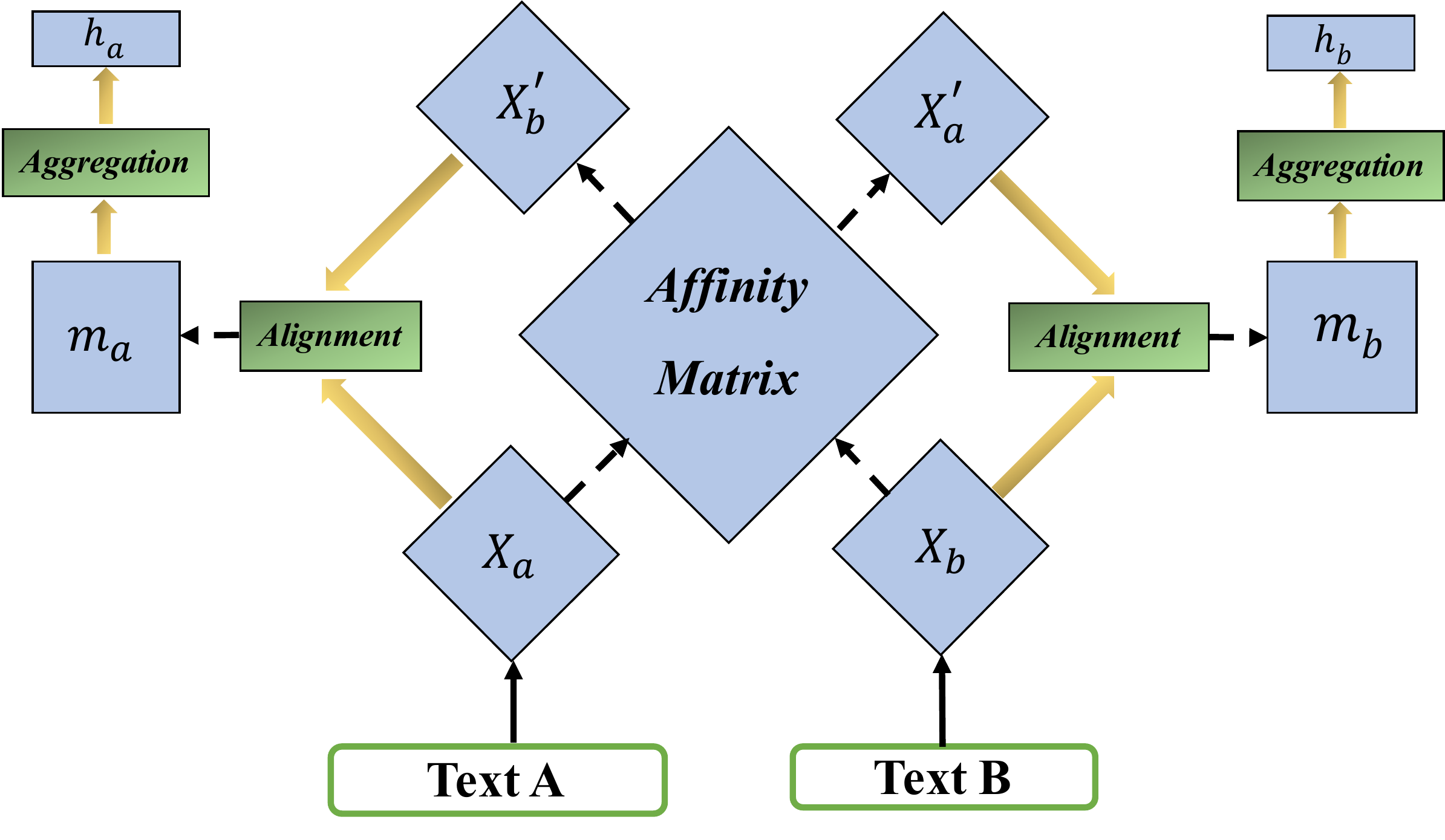}
	\caption{Schematic of the proposed \emph{fine-grained word alignment module} for incorporating textual information into a network embedding. In this setup, word-by-word matching features are explicitly abstracted to infer the relationship between vertices.}
	\label{fig:model}
\end{figure}

\subsection{Fine-Grained Word-by-Word Alignment}
%
With the alignment in the previous section, word-by-context matching features $\alpha_i$ are modeled; however, the \emph{word-by-word} alignment information (\emph{fine-grained}), which is key to characterize the relationship between two vertices (as discussed in the above), is not explicitly captured.
So motivated, we further propose an architecture to explicitly abstract word-by-word alignment information from $\boldsymbol{t}_a$ and $\boldsymbol{t}_b$, to learn the relationship between the two vertices.
This is inspired by the recent success of Relation Networks (RNs) for relational reasoning \cite{santoro2017simple}.

As illustrated in Figure~\ref{fig:model}, given two input embedding matrices $\Xmat_a$ and $\Xmat_b$, we first compute the affinity matrix $\Amat\in\mathbb{R}^{M_b\times M_a}$, whose elements represent the affinity scores corresponding to all word pairs between sequences $\boldsymbol{t}_a$ and $\boldsymbol{t}_b$
\begin{align}\label{eqn:affinity}
	\Amat = \Xmat^T_b\Xmat_a \,.
\end{align}
Subsequently, we compute the context-aware matrix for sequence $\boldsymbol{t}_b$ as
\begin{align}\label{eqn:matrix}
		\Amat_b = \textrm{softmax}(\Amat) \,, \qquad \widetilde{\Xmat}_b = \Xmat_b\Amat_b \,,
\end{align}
where the $\textrm{softmax}(\cdot)$ function is applied column-wise to $\Amat$, and thus $\Amat_b$ contains the attention weights (importance scores) across sequence $\boldsymbol{t}_b$ (columns), which account for each word in sequence $\boldsymbol{t}_a$ (rows).
%
Thus, matrix $\widetilde{\Xmat}_b \in\mathbb{R}^{d\times M_a}$ in \eqref{eqn:matrix} constitutes an attention-weighted embedding for $\Xmat_b$. 
Specifically, the $i$-th column of $\widetilde{\Xmat}_b$, denoted as $\widetilde{\xv}^{(i)}_b$, can be understood as a weighted average over all the words in $\boldsymbol{t}_b$, where higher attention weights indicate better alignment (match) with the $i$-th word in $\boldsymbol{t}_a$.

To abstract the word-by-word alignments, we compare $\xv^{(i)}_a$ with $\widetilde{\xv}^{(i)}_b$, for $i=1,2,...,M_a$, to obtain the corresponding matching vector
\begin{align}\label{eqn:m} 
	\mv^{(i)}_a=f_{\textrm{align}}\left(\xv^{(i)}_a,\widetilde{\xv}^{(i)}_b\right) \,,
\end{align}
where $f_{\textrm{align}}(\cdot)$ represents the alignment function.
Inspired by the observation in \citet{Wang2016ACM} that simple comparison/alignment functions based on element-wise operations exhibit excellent performance in matching text sequences, here we use a combination of element-wise subtraction and multiplication as
%
\begin{align*}
f_{\textrm{align}}(\xv^{(i)}_a,\widetilde{\xv}^{(i)}_a) = [\xv^{(i)}_a - \widetilde{\xv}^{(i)}_a; \xv^{(i)}_a \odot  \widetilde{\xv}^{(i)}_a] \,,
\end{align*}
where $\odot$ denotes the element-wise Hadamard product, then these two operations are concatenated to produce the matching vector $\mv^{(i)}_a$.
Note these operators may be used individually or combined as we will investigate in our experiments.

Subsequently, matching vectors from \eqref{eqn:m} are aggregated to produce the final textual embedding $\hv_t^a$ for sequence $\boldsymbol{t}_a$ as
\begin{align}\label{eqn:hvt}
	\hv_t^a=f_{\textrm{aggregate}}\left(\mv^{(1)}_a,\mv^{(2)}_a,...,\mv^{(M_a)}_a\right) \,,
\end{align}
where $f_{\textrm{aggregate}}$ denotes the aggregation function, which we specify as the max-pooling pooling operation.
Notably, other commutative operators, such as summation or average pooling, can be otherwise employed.
Although these aggregation functions are simple and invariant to the order of words in input sentences, they have been demonstrated to be highly effective in relational reasoning \cite{parikh2016decomposable, santoro2017simple}.
To further explore this, in Section~\ref{sec:ablation}, we conduct an ablation study comparing different choices of alignment and aggregation functions.

The representation $\hv_b$ can be obtained in a similar manner through \eqref{eqn:affinity}, \eqref{eqn:matrix}, \eqref{eqn:m} and \eqref{eqn:hvt}, but replacing \eqref{eqn:affinity} with $\Amat = \Xmat^T_a\Xmat_b$ (its transpose).
Note that this word-by-word alignment is more computationally involved than word-by-context; however, the former has substantially fewer parameters to learn, provided we no longer have to estimate the parameters in \eqref{eqn:alpha}. 

\subsection{Training and Inference}
For large-scale networks, computing and optimizing the conditional probabilities in \eqref{eq:obj} using \eqref{eq:def} is computationally prohibitive, since it requires the summation over all vertices $\boldsymbol{V}$ in $\boldsymbol{G}$.
To address this limitation, we leverage the negative sampling strategy introduced by \citet{mikolov2013distributed}, \emph{i.e.}, we perform computations by sampling a subset of negative edges.
As a result, the conditional in \eqref{eq:def} can be rewritten as:
\begin{align*}
	\begin{aligned}
	p(\hv^i|\hv^j) & = \log \sigma\left({\hv^j}^T \hv^i\right) \\
	& + \sum_{i=1}^{K} \mathbb{E}_{\hv^i\sim P(v)}\left[\log \sigma\middle(-{\hv^j}^T \hv^i\middle)\right] \,,
	\end{aligned}
\end{align*}
where $\sigma(x) = 1/(1+\exp(-x))$ is the sigmoid function.
Following \citet{mikolov2013distributed}, we set the noise distribution $P(v) \propto d_v^{3/4}$, where $d_v$ is the out-degree of vertex $v\in\boldsymbol{V}$.
The number of negative samples $K$ is treated as a hyperparameter.
We use Adam \cite{kingma2014adam} to update the model parameters
while minimizing the objective in \eqref{eq:obj}.

\section{Related Work}
Network embedding
methods can be divided into two categories: 
(\emph{\romannumeral1}) methods that solely rely on the structure, \emph{e.g.}, vertex information; and (\emph{\romannumeral2}) methods that leverage both the structure the network and the information associated with its vertices.

For the first type of models, DeepWalk \cite{perozzi2014deepwalk} has been proposed to learn node representations by generating node contexts via truncated random walks; it is similar to the concept of Skip-Gram \cite{mikolov2013distributed}, originally introduced for learning word embeddings.
LINE \cite{tang2015line} proposed a principled objective to explicitly capture first-order and second-order proximity information from the vertices of a network.
Further, \citet{grover2016node2vec} introduced a biased random walk procedure to generate the neighborhood for a vertex, which infers the node representations by maximizing the likelihood of preserving the local context information of vertices.
However, these algorithms generally ignore rich heterogeneous information associated with vertices.
Here, we focus on incorporating \emph{textual information} into network embeddings.

To learn semantic-aware network embeddings, Text-Associated DeepWalk (TADW) \cite{yang2015network} proposed to integrate textual features into network representations with matrix factorization, by leveraging the equivalence between DeepWalk and matrix factorization.
CENE (Content-Enhanced Network Embedding) \cite{sun2016general} used bidirectional recurrent neural networks to abstract the semantic information associated with vertices, which further demonstrated the advantages of employing textual information.
To capture the interaction between sentences of vertex pairs, \citet{tu2017cane} further proposed Context-Aware Network Embedding (CANE), that employs a mutual attention mechanism to adaptively account for the textual information from neighboring vertices.
Despite showing improvement over structure-only models, these semantic-aware methods cannot capture word-level alignment information, which is important for inferring the relationship between node pairs, as previously discussed.
In this work, we introduce a Word-Alignment-based Network Embedding (WANE) framework, which aligns and aggregates word-by-word matching features in an explicit manner, to obtain more informative network representations.

\section{Experimental Setup}

\paragraph{Datasets}
We investigate the effectiveness of the proposed WANE model on two standard network-embedding-based tasks, \emph{i.e.}, link prediction and multi-label vertex classification.
The following three real-world datasets are employed for quantitative evaluation: 
(\emph{\romannumeral1}) \emph{Cora}, a standard paper citation network that contains 2,277 machine learning papers (vertices) grouped into 7 categories and connected by 5,214 citations (edges)
(\emph{\romannumeral2}) \emph{HepTh}, another citation network of 1,038 papers with abstract information and 1,990 citations;
(\emph{\romannumeral3}) \emph{Zhihu}, a network of 10,000 active users from Zhihu, the largest Q\&A website in China, where 43,894 vertices and descriptions of the Q\&A topics are available.
The average lengths of the text in the three datasets are 90, 54, and 190, respectively.
To make direct comparison with existing work, we employed the same preprocessing procedure\footnote{https://github.com/thunlp/CANE} of \citet{tu2017cane}.

\begin{table*}  \small 
	\centering
	\begin{tabular}{c||c|c|c|c|c|c|c|c|c}
		\toprule[1.2pt]
		\textbf{\%Training Edges} &  	\textbf{15\%} & \textbf{25\%} & 	\textbf{35\%} & \textbf{45\%} & \textbf{55\%} & \textbf{65\%} & \textbf{75\%} & \textbf{85\%} & \textbf{95\%} \\
		\hline
		\textbf{MMB}      & 54.7 & 57.1 & 59.5 & 61.9 & 64.9 & 67.8 & 71.1 & 72.6 & 75.9  \\
		\textbf{DeepWalk}    & 56.0 & 63.0 & 70.2 & 75.5 & 80.1 & 85.2 & 85.3 & 87.8 & 90.3   \\ 
		\textbf{LINE}           & 55.0 & 58.6 & 66.4 & 73.0 & 77.6 & 82.8 & 85.6& 88.4& 89.3  \\ 
		\textbf{node2vec}        & 55.9 & 62.4 & 66.1 & 75.0 & 78.7 & 81.6 & 85.9 & 87.3 & 88.2   \\
		\hline 
		\textbf{Naive combination}       & 72.7 & 82.0 & 84.9 & 87.0 & 88.7 & 91.9 & 92.4 & 93.9 & 94.0  \\ 
		\textbf{TADW}            & 86.6 & 88.2 & 90.2 & 90.8 & 90.0 & 93.0 & 91.0 & 93.4 & 92.7  \\ 
		\textbf{CENE} & 72.1 & 86.5 &84.6& 88.1& 89.4 &89.2& 93.9 &95.0 &95.9  \\
		\textbf{CANE}  & 86.8& 91.5 &92.2 &93.9 &94.6 &94.9 &95.6 &96.6& 97.7  \\
		\hline
		\textbf{WANE}  & 86.1 & 90.9 & 92.3 & 93.1 & 93.4 & 94.5 & 95.1 & 95.4 & 95.9 \\
		\textbf{WANE-\emph{wc}}  & 88.7 & 92.1 & 92.9 &94.4 & 94.8 & 95.1 & 95.7 & 96.5 & 97.4 \\
		\textbf{WANE-\emph{ww}}  & \textbf{91.7} & \textbf{93.3} & \textbf{94.1} & \textbf{95.7} & \textbf{96.2} & \textbf{96.9} & \textbf{97.5} & \textbf{98.2} & \textbf{99.1}  \\
		\bottomrule[1.2pt]
	\end{tabular}
	\caption{AUC scores for link prediction on the \emph{Cora} dataset.}
	\label{tab:cora}
\end{table*}

\begin{table*}[!ht]  \small 
	\centering
	\begin{tabular}{c||c|c|c|c|c|c|c|c|c}
		\toprule[1.2pt]
		\textbf{\%Training Edges} &  	\textbf{15\%} & \textbf{25\%} & 	\textbf{35\%} & \textbf{45\%} & \textbf{55\%} & \textbf{65\%} & \textbf{75\%} & \textbf{85\%} & \textbf{95\%} \\
		\hline
		\textbf{MMB}      &54.6& 57.9 &57.3 &61.6& 66.2 &68.4 &73.6 &76.0 &80.3  \\
		\textbf{DeepWalk}    & 55.2& 66.0 &70.0& 75.7& 81.3 &83.3 &87.6 &88.9 &88.0   \\ 
		\textbf{LINE}            & 53.7 &60.4 &66.5 &73.9 &78.5 &83.8 &87.5 &87.7 &87.6  \\ 
		\textbf{node2vec}        & 57.1 &63.6 &69.9 &76.2 &84.3 &87.3& 88.4 &89.2 &89.2  \\
		\hline 
		\textbf{Naive combination}       & 78.7 &82.1& 84.7& 88.7 &88.7 &91.8 &92.1 &92.0 &92.7  \\ 
		\textbf{TADW}            & 87.0 &89.5 &91.8& 90.8 &91.1 &92.6 &93.5 &91.9 &91.7  \\
		\textbf{CENE} & 86.2 &84.6 &89.8 &91.2 &92.3 &91.8 &93.2 &92.9 &93.2  \\
		\textbf{CANE}  & 90.0 & 91.2 &92.0 &93.0 &94.2 &94.6 &95.4 &95.7 &96.3  \\
		\hline
		\textbf{WANE}  & 88.5 & 90.7 & 91.1 & 92.6 & 93.5 & 94.2 & 94.9 & 95.3 & 95.8 \\
		\textbf{WANE-\emph{wc}}  & 90.1 & 91.4 & 91.9 &94.1 & 95.3 & 95.9 & 96.5 & 96.9 & 97.2 \\
		\textbf{WANE-\emph{ww}}  & \textbf{92.3} & \textbf{94.1} & \textbf{95.7} & \textbf{96.7} & \textbf{97.5} & \textbf{97.5} & \textbf{97.7} & \textbf{98.2} & \textbf{98.7}  \\
		\bottomrule[1.2pt]
	\end{tabular}
	\caption{AUC scores for link prediction on the \emph{HepTh} dataset.}
	\label{tab:hepth}
	\vspace{-2mm}
\end{table*}

\paragraph{Training Details}
For fair comparison with CANE \cite{tu2017cane}, we set the dimension of network embedding for our model to 200.
The number of negative samples $K$ is selected from $\{1, 3, 5\}$ according to performance on the validation set. 
We set the batch size as 128, and the model is trained using Adam \cite{kingma2014adam}, with a learning rate of $1 \times 10^{-3}$ for all parameters.
Dropout regularization is employed on the word embedding layer, with rate selected from $\{0.5, 0.7, 0.9\}$, also on the validation set.
Our code will be released to encourage future research.

\paragraph{Baselines}
To evaluate the effectiveness of our framework, we consider several strong baseline methods for comparisons, which can be categorized into two types: (\emph{\romannumeral1}) models that only exploit \emph{structural} information: MMB \cite{Airoldi2008MixedMS}, DeepWalk \cite{perozzi2014deepwalk}, LINE \cite{tang2015line}, and node2vec \cite{grover2016node2vec}.
(\emph{\romannumeral2}) Models that take both \emph{structural} and \emph{textual} information into account: Naive combination (which simply concatenates the structure-based embedding with CNN-based text embeddings, as explored in \cite{tu2017cane}, TADW \cite{yang2015network}, CENE \cite{sun2016general}, and CANE \cite{tu2017cane}.
It is worth noting that unlike all these baselines, WANE explicitly captures word-by-word interactions between text sequence pairs.

\paragraph{Evaluation Metrics}
We employ AUC \cite{hanley1982meaning} as the evaluation metric for link prediction, which measures the probability that vertices within an existing edge, randomly sampled from the test set, are more similar than those from a random pair of non-existing vertices, in terms of the inner product between their corresponding embeddings.

For multi-label vertex classification and to ensure fair comparison, we follow \citet{yang2015network} and employ a linear SVM on top of the learned network representations, and evaluate classification accuracy with different training ratios (varying from $10\%$ to $50\%$).
The experiments for each setting are repeated 10 times and the average test accuracy is reported.
%

\section{Experimental Results}
We experiment with three variants for our WANE model: (\emph{\romannumeral1}) WANE: where the word embeddings of each text sequence are simply average to obtain the sentence representations, similar to \cite{joulin2016bag, Shen2018Baseline}.
(\emph{\romannumeral2}) WANE-\emph{wc}: where the textual embeddings are inferred with word-by-context alignment.
(\emph{\romannumeral3}) WANE-\emph{ww}: where the word-by-word alignment mechanism is leveraged to capture word-by-word matching features between available sequence pairs.

\begin{table*} [!ht]  \small 
	\centering
	\begin{tabular}{c||c|c|c|c|c|c|c|c|c}
		\toprule[1.2pt]
		\textbf{\%Training Edges} &  	\textbf{15\%} & \textbf{25\%} & 	\textbf{35\%} & \textbf{45\%} & \textbf{55\%} & \textbf{65\%} & \textbf{75\%} & \textbf{85\%} & \textbf{95\%} \\
		\hline
		\textbf{MMB}      & 51.0 & 51.5 & 53.7 & 58.6 & 61.6 & 66.1 & 68.8 & 68.9 & 72.4  \\
		\textbf{DeepWalk}    & 56.6 & 58.1 & 60.1 & 60.0 & 61.8 & 61.9 & 63.3 & 63.7 & 67.8   \\ 
		\textbf{LINE}            & 52.3 & 55.9 & 59.9 & 60.9 & 64.3 & 66.0 & 67.7 & 69.3 & 71.1  \\ 
		\textbf{node2vec}        & 54.2 & 57.1 & 57.3 & 58.3 & 58.7 & 62.5 & 66.2 & 67.6 & 68.5  \\
		\hline 
		\textbf{Naive combination}       & 55.1 & 56.7 & 58.9 & 62.6 & 64.4 & 68.7 & 68.9 & 69.0 & 71.5  \\ 
		\textbf{TADW}            & 52.3 & 54.2 & 55.6 & 57.3 & 60.8 & 62.4 & 65.2 & 63.8 & 69.0  \\
		\textbf{CENE} & 56.2 & 57.4 & 60.3 & 63.0 & 66.3 & 66.0 & 70.2 & 69.8 & 73.8 \\
		\textbf{CANE}  & 56.8 & 59.3 & 62.9 & 64.5 & 68.9 & 70.4 & 71.4 & 73.6 & 75.4  \\
		\hline
		\textbf{WANE}  & 52.1 & 56.6 & 60.7 & 64.2 & 67.5 & 69.1 & 71.3 & 72.8 & 73.9 \\
		\textbf{WANE-\emph{wc}}  & 55.2 & 59.9 & 64.2 & 68.1 & 71.3 & 73.4 & 75.6 & 76.3 &78.8 \\
		\textbf{WANE-\emph{ww}}  & \textbf{58.7} & \textbf{63.5} & \textbf{68.3} & \textbf{71.9} & \textbf{74.9} & \textbf{77.0} & \textbf{79.7} & \textbf{80.0} & \textbf{82.6}  \\
		\bottomrule[1.2pt]
	\end{tabular}
	\caption{AUC scores for link prediction on the \emph{Zhihu} dataset.}
	\label{tab:zhihu}
\end{table*}

\begin{figure*}[!ht] \centering
	\begin{tabular}{ccc}  
		\includegraphics[width=4.0cm]{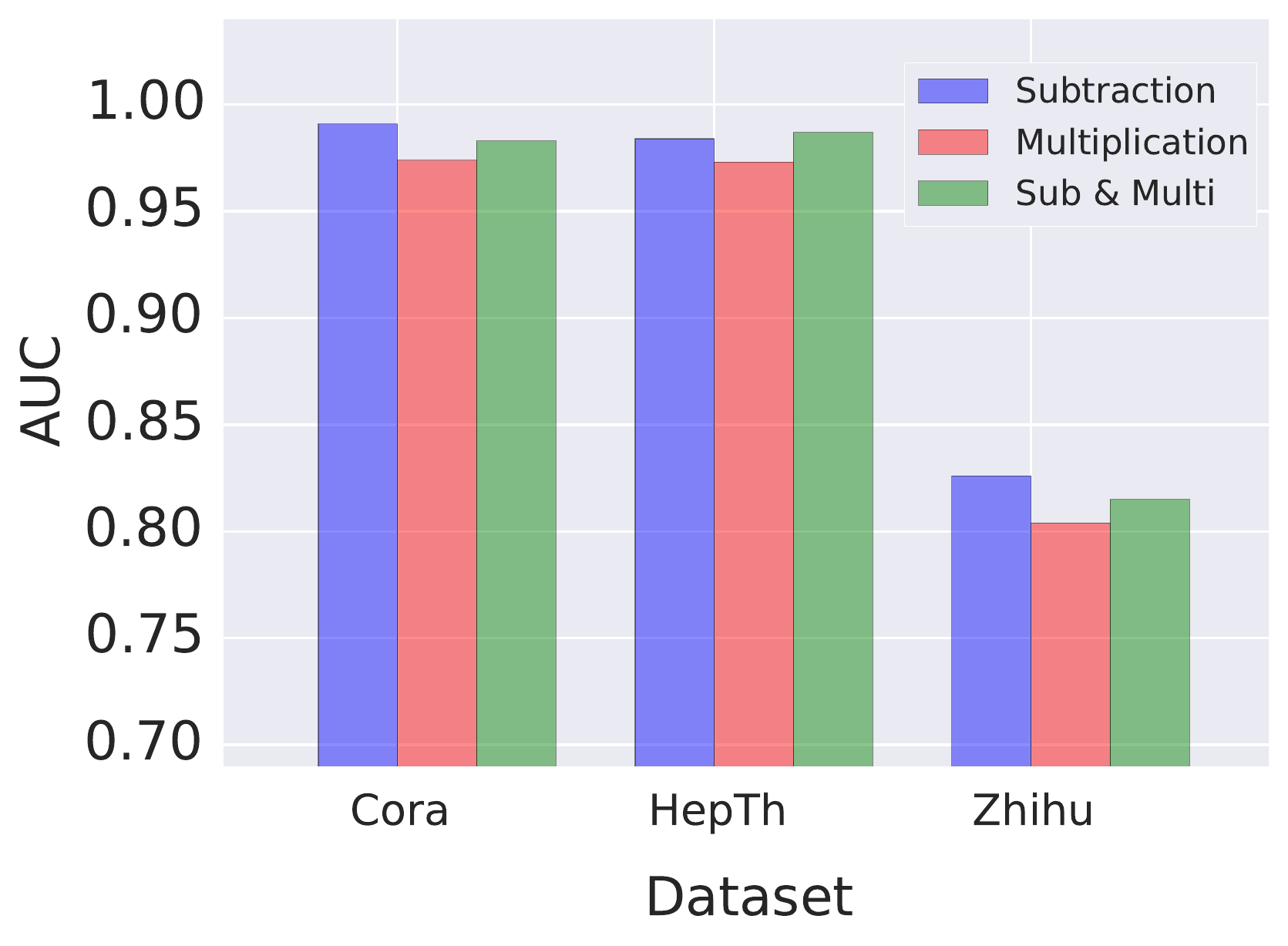} & \hspace{2mm}
		\includegraphics[width=4.0cm]{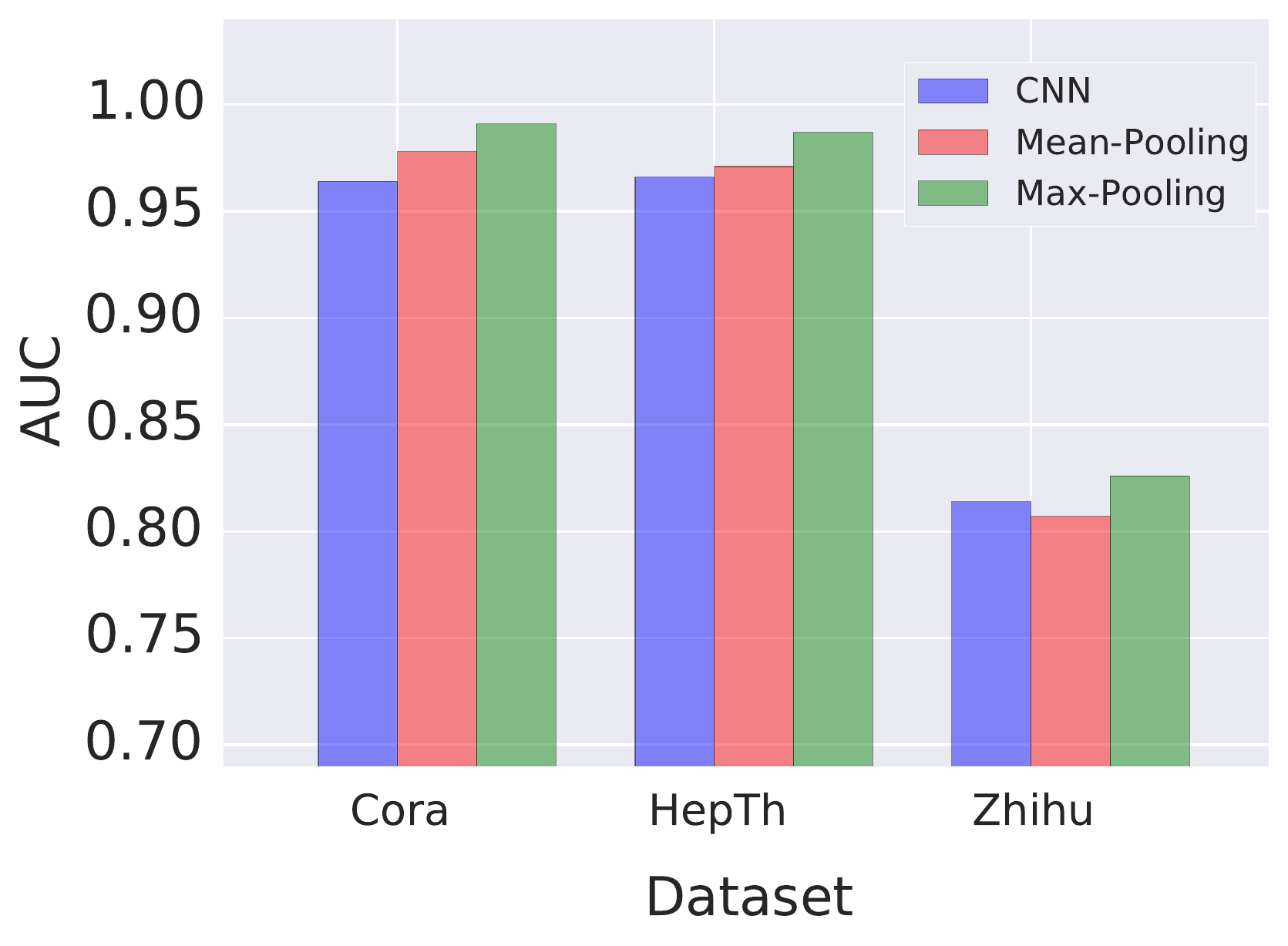} &  \hspace{ 2mm}
		\includegraphics[width=4.0cm]{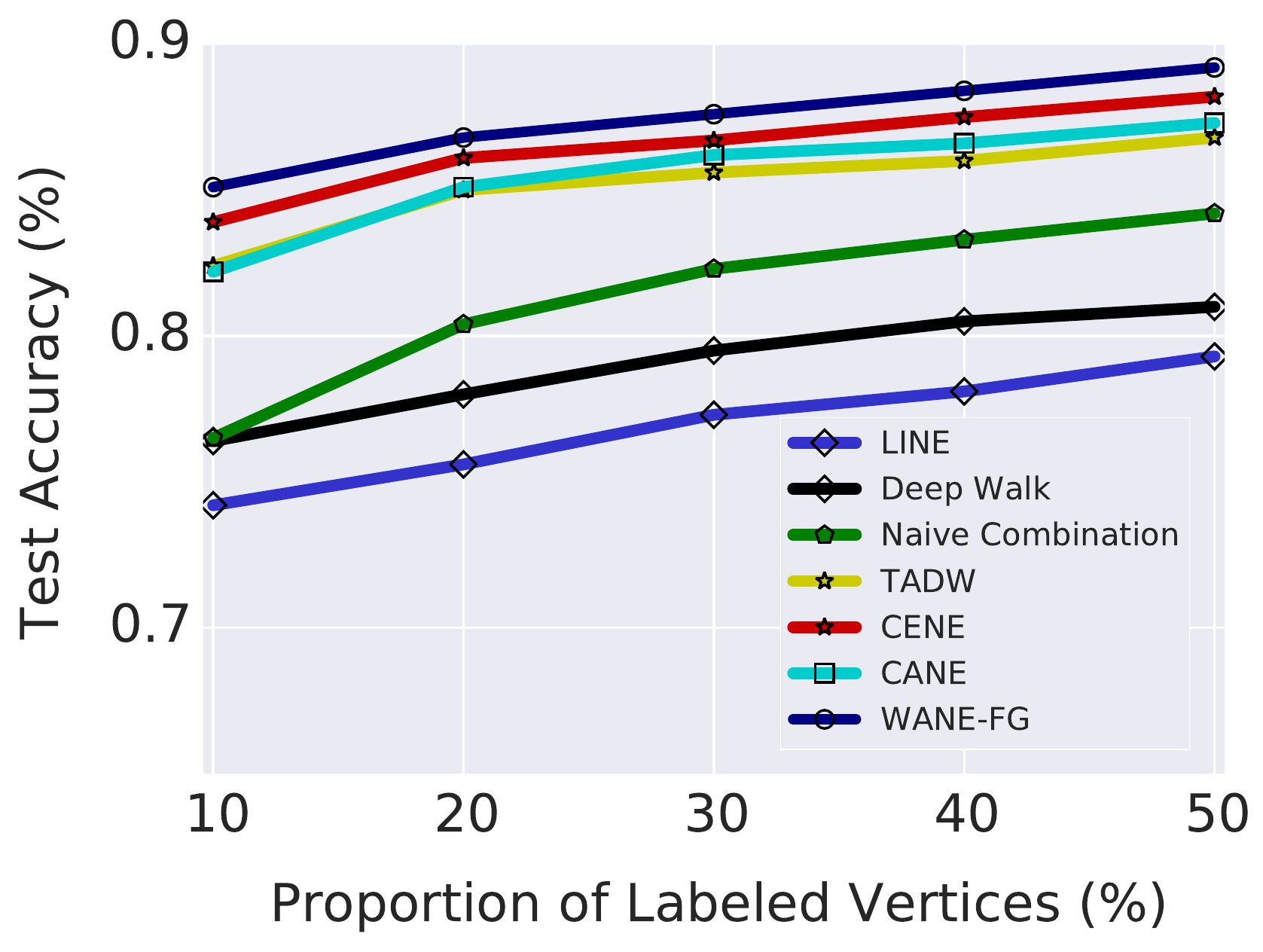} 
		\\
		\hspace{-0mm}
		(a) $f_{\textrm{align}}$ \vspace{0mm}  & 
		\hspace{-0mm}
		(b) $f_{\textrm{aggregate}}$ \hspace{-0mm}& 
		(c) vertex classification \hspace{-0mm}
	\end{tabular}
	\vspace{-2mm}
	\caption{(a, b) Ablation study on the choice of different alignment and aggregation functions. (c) Test accuracy of \emph{supervised} vertex classification on the \emph{Cora} dataset.}
	\label{fig:classification}
	\vspace{-3mm}
\end{figure*}

\subsection{Link Prediction}
Table~\ref{tab:cora} presents link prediction results for all models on Cora dataset, where different ratios of edges are used for training.
It can be observed that when only a small number of edges are available, \emph{e.g.}, $15\%$, the performances of structure-only methods is much worse than semantic-aware models that have taken textual information into consideration 
The perfromance gap tends to be smaller when a larger proportion of edges are employed for training.
This highlights the importance of incorporating associated text sequences into network embeddings, especially in the case of representing a relatively sparse network.
More importantly, the proposed WANE-\emph{ww} model consistently outperforms other semantic-aware NE models by a substantial margin, indicating that our model better abstracts word-by-word alignment features from the text sequences available, thus yields more informative network representations.

Further, WANE-\emph{ww} also outperforms WANE or WANE-\emph{wc} on a wide range of edge training proportions.
This suggests that: 
(\emph{\romannumeral1}) adaptively assigning different weights to each word within a text sequence (according to its paired sequence) tends to be a better strategy than treating each word equally (as in WANE).
(\emph{\romannumeral2}) Solely considering the context-by-word alignment features (as in WANE-\emph{wc}) is not as efficient as abstracting word-by-word matching information from text sequences.
We observe the same trend and the superiority of our WANE-\emph{ww} models on the other two datasets, HepTh and Zhihu datasets, as shown in Table~\ref{tab:hepth} and \ref{tab:zhihu}, respectively.

\subsection{Multi-label Vertex Classification}
We further evaluate the effectiveness of proposed framework on vertex classification tasks with the Cora dataset.
Similar to \citet{tu2017cane}, we generate the global embedding for each vertex by taking the average over its \emph{context-aware} embeddings with all other connected vertices.
As shown in Figure~\ref{fig:classification}(c), semantic-aware NE methods (including naive combination, TADW, CENE, CANE) exhibit higher test accuracies than semantic-agnostic models, demonstrating the advantages of incorporating textual information. 
Moreover, WANE-\emph{ww} consistently outperforms other competitive semantic-aware models on a wide range of labeled proportions, suggesting that explicitly capturing word-by-word alignment features is not only useful for vertex-pair-based tasks, such as link prediction, but also results in better global embeddings which are required for vertex classification tasks.
These observations further demonstrate that WANE-\emph{ww} is an effective and robust framework to extract informative network representations.

\paragraph{Semi-supervised classification}
We further consider the case where the training ratio is less than $10\%$, and evaluate the learned network embedding with a semi-supervised classifier.
Following \citet{yang2015network}, we employ a Transductive SVM (TSVM) classifier with a linear kernel \cite{joachims1998making} for fairness.
As illustrated in Table~\ref{tab:semi}, the proposed WANE-\emph{ww} model exhibits superior performances in most cases.
This may be due to the fact that WANE-\emph{ww} extracts information from the vertices and text sequences jointly, thus the obtained vertex embeddings are less noisy and perform more consistently with relatively small training ratios \cite{yang2015network}.

\subsection{Ablation Study} \label{sec:ablation}
Motivated by the observation in \citet{Wang2016ACM} that the advantages of different functions to match two vectors vary from task to task, we further explore the choice of alignment and aggregation functions in our WANE-\emph{ww} model.
To match the word pairs between two sequences, we experimented with three types of operations: \emph{subtraction}, \emph{multiplication}, and \emph{Sub \& Multi} (the concatenation of both approaches).
As shown in Figure~\ref{fig:classification}(a) and \ref{fig:classification}(b), element-wise \emph{subtraction} tends to be the most effective operation performance-wise on both Cora and Zhihu datasets, and performs comparably to \emph{Sub \& Multi} on the HepTh dataset.
This finding is consistent with the results in \citet{Wang2016ACM}, where they found that simple comparison functions based on \emph{element-wise} operations work very well on matching text sequences.

In terms of the aggregation functions, we compare (one-layer) \emph{CNN}, \emph{mean-pooling}, and \emph{max-pooling} operations to accumulate the matching vectors.
As shown in Figure~\ref{fig:classification}(b), \emph{max-pooling} has the best empirical results on all three datasets. 
This may be attributed to the fact that the \emph{max-pooling} operation is better at selecting important word-by-word alignment features, among all matching vectors available, to infer the relationship between vertices.

\subsection{Qualitative Analysis}
\paragraph{Embedding visualization}
To visualize the learned network representations, we further employ $t$-SNE to map the low-dimensional vectors of the vertices to a 2-D embedding space.
We use the \emph{Cora} dataset because there are labels associated with each vertex and WANE-\emph{ww} to obtain the network embeddings.

\begin{figure}
	\centering
	\vspace{-4mm}
	\includegraphics[width=.38\textwidth]{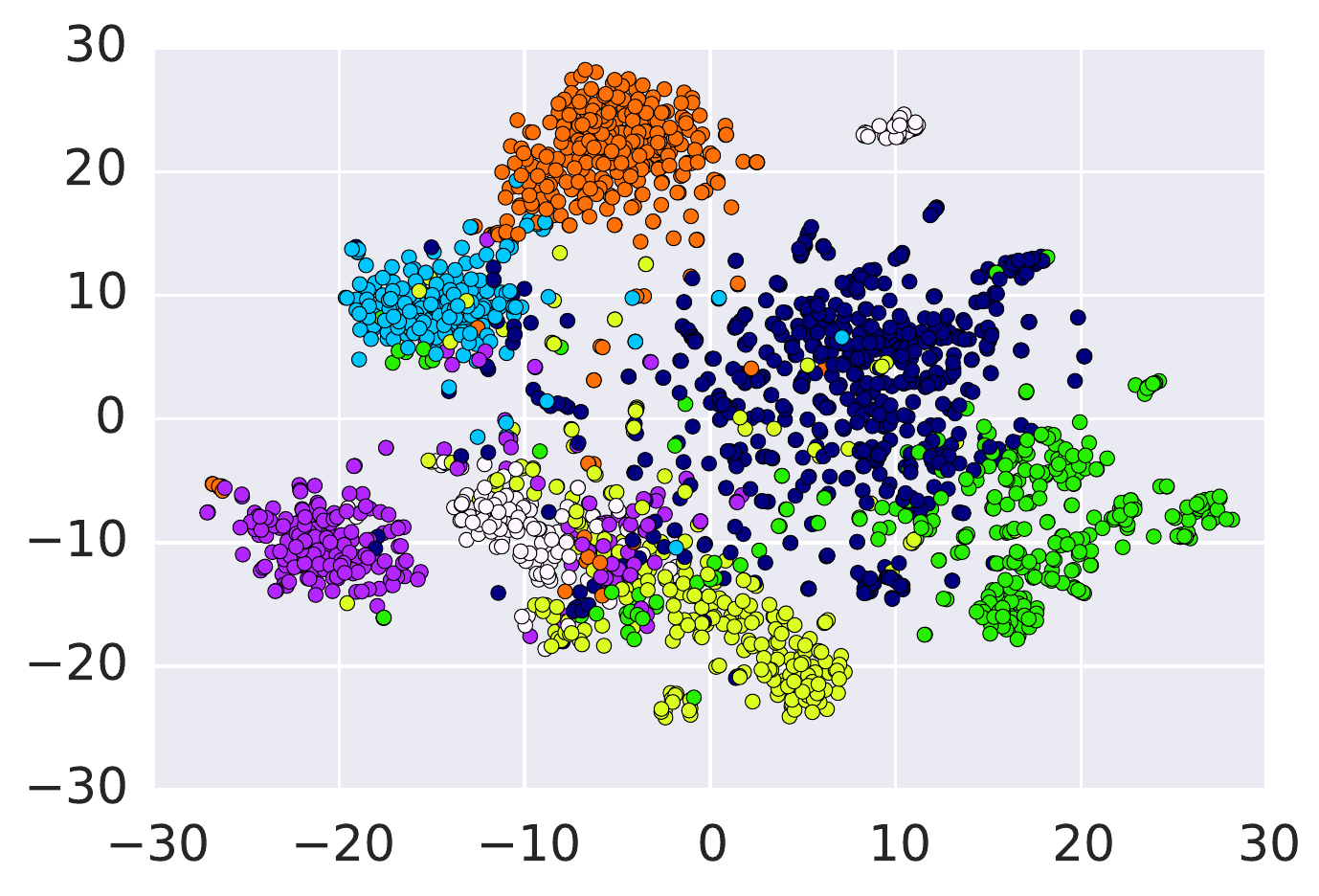}
	\vspace{-2mm}
	\caption{$t$-SNE visualization of the learned network embeddings on the Cora dataset.}
	\vspace{-4mm}
	\label{fig:tsne}
\end{figure}

As shown in Figure~\ref{fig:tsne} where each point indicates one paper (vertex), and the color of each point indicates the \emph{category} it belongs to, the embeddings of the same label are indeed very close in the 2-D plot, while those with different labels are relatively farther from each other.
Note that the model is not trained with any label information, indicating that WANE-\emph{ww} has extracted meaningful patterns from the text and vertex information available.

\begin{table}
	\centering
	\small
	\begin{tabular}{c||c|c|c|c}
		\textbf{Baseline Models} &  1\% & 3\% & 7\% & 10\%   \\
		\hline
		\textbf{Text Only} & 33.0 & 43.0 & 57.1 & 62.8 \\
		\textbf{Naive Combination}   & 67.4 & 70.6 & 75.1 & 77.4    \\ 
		\textbf{TADW}   & 72.1 & 77.0 & 79.1 & 81.3    \\ 
		\textbf{CENE}  & \textbf{73.8} & 79.1 & 81.5 & 84.5     \\ 
		\textbf{CANE}  & 72.6 & 78.2 & 80.4 & 83.4    \\
		\hline
		\textbf{WANE-\emph{ww}} (ours)    & 73.4 & \textbf{79.6} & \textbf{82.7} & \textbf{85.1}     \\ 
	\end{tabular}
	\vspace{-2mm}
	\caption{\emph{Semi-supervised} vertex classification results on the \emph{Cora} dataset.}
	\vspace{-4mm}
	\label{tab:semi}
\end{table}

\begin{figure}[!h]
	\centering
	\begin{tabular}{c}
	\hspace{-3mm}
	\includegraphics[width=7.6cm]{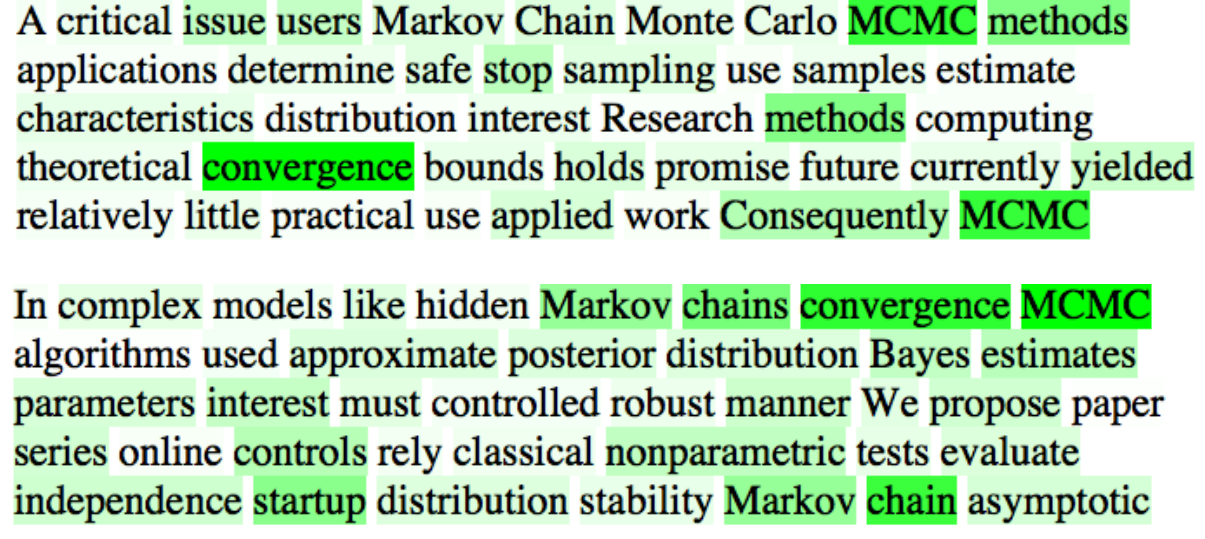} \vspace{-2mm} \\
	\small {(a) Sentence pairs associated with Edge \#1}  \vspace{2mm}  \\
	\hspace{-3mm}
	\includegraphics[width=7.6cm]{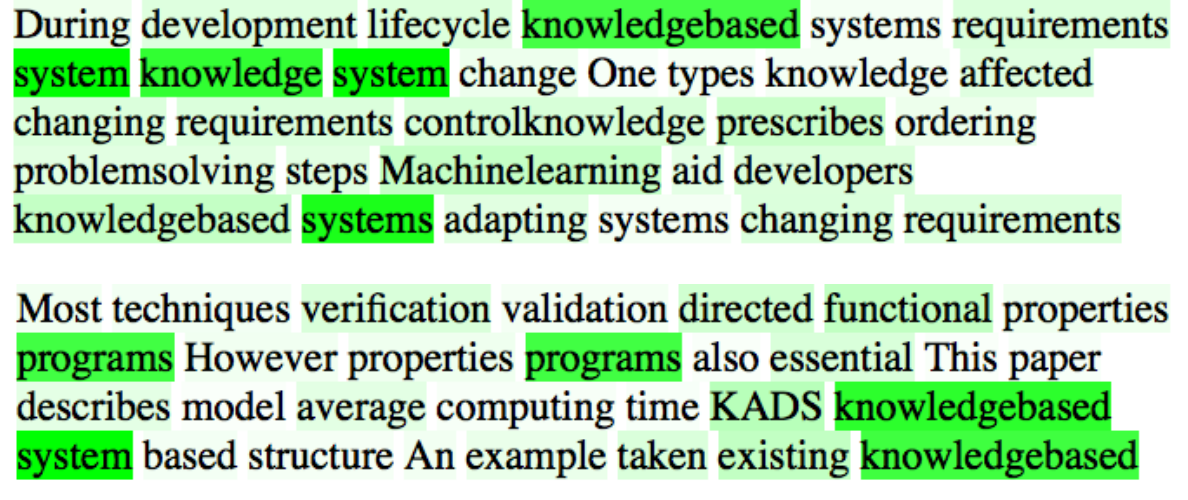} \vspace{-1mm}  \\
	\small {(b) Sentence pairs associated with Edge \#2}   \hspace{-0mm}
	\end{tabular}
	\caption{Visualization of the word-level \emph{matching vectors}. Darker shades represent larger values of the norm of $\mv^{(i)}$ at each word position.}
	\label{fig:attention}
	\vspace{-4mm}
\end{figure}

\paragraph{Case study}
The proposed word-by-word alignment mechanism can be used to highlight the most informative
words (and the corresponding matching features) wrt the relationship between vertices.
We visualize the norm of matching vector obtained in \eqref{eqn:m} in Figure~\ref{fig:attention} for the Cora dataset.
It can be observed that matched key words, \emph{e.g.}, `MCMC', `convergence', between the text sequences are indeed assigned higher values in the matching vectors.
These words would be selected preferentially by the final \emph{max-pooling} aggregation operation.
This indicates that WANE-\emph{ww} is able to abstract important word-by-word alignment features from paired text sequences.

\section{Conclusions}
We have presented a novel framework to incorporate the semantic information from vertex-associated text sequences into network embeddings.
An \emph{align-aggregate} framework is introduced, which first aligns a sentence pair by capturing the word-by-word matching features, and then adaptively aggregating these word-level alignment information with an efficient \emph{max-pooling} function.
The semantic features abstracted are further encoded, along with the structural information, into a shared space to obtain the final network embedding.
Compelling experimental results on several tasks demonstrated the advantages of our approach.
In future work, we aim to leverage abundant unlabeled text data to abstract more informative sentence representations \cite{dai2015semi, zhang2017deconvolutional,  shen2017deconvolutional, tang2018multi} .
Another interesting direction is to learn \emph{binary} and compact network embedding, which could be more efficient in terms of both computation and memory, relative to its continuous counterpart \citep{shen2018nash}.

\paragraph{Acknowledgments}
This research was supported in part by DARPA, DOE, NIH, ONR and NSF.


\bibliography{ne}
\bibliographystyle{acl_natbib}

\end{document}